\pdfoutput=1
\documentclass[11pt]{article}

\usepackage[final]{acl}

\usepackage{times}
\usepackage{latexsym}
\usepackage{bm}
\usepackage{amsmath}
\usepackage{amsfonts}
\usepackage{algorithm, algpseudocode}
\usepackage[caption=false,font=footnotesize,labelfont=rm,textfont=rm]{subfig}
\usepackage{multirow}
\usepackage{booktabs}
\usepackage{enumitem}
\usepackage[T1]{fontenc}
\usepackage[utf8]{inputenc}

\usepackage{microtype}

\usepackage{inconsolata}

\usepackage{graphicx}

\title{IAM: Efficient Inference through Attention Mapping between Different-scale LLMs}

\author{Yi Zhao$^{1,2,3}$, Zuchao Li$^{4,*}$ and Hai Zhao$^{1,2,3,}$\thanks{$\ $  Corresponding author. This research was supported by the Joint Research Project of
Yangtze River Delta Science and Technology Innovation Community (No.
2022CSJGG1400), the National Natural Science Foundation of China (No. 62306216) and Xiaomi Open-Competition Research Program.} \\
$^{1}$School of Computer Science, Shanghai Jiao Tong University\\
$^{2}$Key Laboratory of Shanghai Education Commission for Intelligent Interaction\\ and Cognitive Engineering, Shanghai Jiao Tong University\\
$^{3}$Shanghai Key Laboratory of Trusted Data Circulation and Governance in Web3\\
$^{4}$School of Artificial Intelligence, Wuhan University, Wuhan, China \\
{\tt zhao-yi@sjtu.edu.cn, zcli-charlie@whu.edu.cn,}\\
{\tt zhaohai@cs.sjtu.edu.cn}\\
}

\begin{document}
\maketitle
\begin{abstract}

LLMs encounter significant challenges in resource consumption nowadays, especially with long contexts. Despite extensive efforts dedicate to enhancing inference efficiency,  these methods primarily exploit internal sparsity within the models, without leveraging external information for optimization. We identify the high similarity of attention matrices across different-scale LLMs, which offers a novel perspective for optimization. We first conduct a comprehensive analysis of how to measure similarity, how to select mapping Layers and whether mapping is consistency. Based on these insights, we introduce the IAM framework, which achieves dual benefits of accelerated attention computation and reduced KV cache usage by performing attention mapping between small and large LLMs. Our experimental results demonstrate that IAM can accelerate prefill by 15\% and reduce KV cache usage by 22.1\% without appreciably sacrificing performance. Experiments on different series of models show the generalizability of IAM. Importantly, it is also orthogonal to many existing KV cache optimization methods, making it a versatile addition to the current toolkit for enhancing LLM efficiency. Our code is available at \href{https://github.com/QQQ-yi/IAM}{https://github.com/QQQ-yi/IAM}

\iffalse
With the advent of assistive technologies such as Retrieval-Augmented Generation (RAG) and the emergence of reasoning models, large language models (LLMs) face significant challenges due to the rapid expansion of KV caches in handling long contexts. While numerous efforts have been dedicated to KV cache compression, these methods primarily exploit internal sparsity and similarity within the models, without leveraging external information for optimization. We identify a high similarity of attention matrices across different-scale models with same series, which offers a novel perspective for optimizing KV cache usage. We first conduct a comprehensive analysis of choosing similarity metrics, the characteristics of different mapped layers, and the consistency of mapping relations. Based on these insights, we introduce the IAM framework, which achieves dual benefits of reduced KV cache usage and accelerated attention computation by performing attention mapping between small and large language models. Our experimental results demonstrate that IAM can reduce KV cache usage by 22.1\% and achieve an 11\% improvement in throughput without sacrificing performance. Experiments on different series of models confirm the generalizability of IAM. Importantly, it is also orthogonal to many existing KV cache optimization methods, making it a versatile addition to the current toolkit for enhancing LLM efficiency.
\fi
\end{abstract}

\section{Introduction}
Large language models (LLMs) like GPT4 \cite{openai2024gpt4technicalreport} have emerged with remarkable natural language understanding capabilities and broad prospects in application. While subsequent advancements such as Chain-of-Thought (CoT) \cite{wei2022chain,yao2024tree} have revitalized the landscape of applications, they also introduce significant computation and memory consumption due to exceedingly long contexts. Recent developments in reasoning models, exemplified by ChatGPT-o1 \cite{chatgpt-o1} and DeepSeek-R1 \cite{deepseekai2025deepseekr1incentivizingreasoningcapability}, have exacerbated this issue because of their extensive internal reasoning processes.

%Subsequent advancements such as In-Context Learning (ICL) \cite{brown2020language,dong2024survey}, Chain-of-Thought (CoT) \cite{wei2022chain,yao2024tree} and Retrieval Augmented Generation (RAG) \cite{lewis2020retrieval} have significantly revitalized the landscape of applications based on LLMs. These technologies expand the capabilities of LLMs by enabling the activation of domain-specific knowledge or strengthening memory capabilities. However, they also introduce significant computation and memory consumption due to exceedingly long contexts. 

\begin{figure}[!htb]
\centering
\includegraphics[scale=0.13]{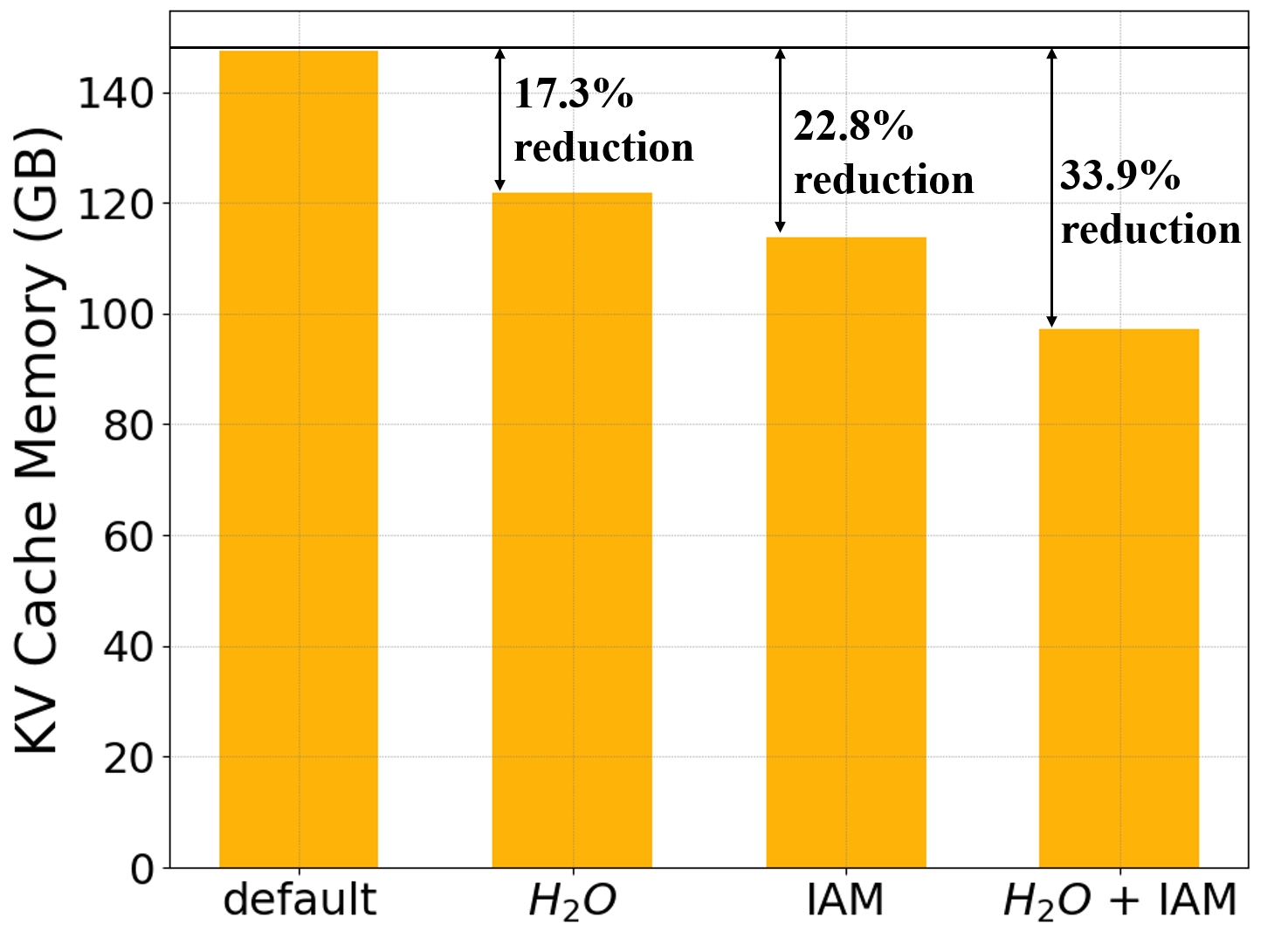}
\caption{IAM is orthogonal to the existing KV cache eviction methods and can further reduce the KV cache usage in long context scenarios.}
\label{fig:reduce}
\end{figure}

To address the aforementioned challenges and achieve efficient inference, various methodologies have been proposed, including optimizing model architectures \cite{sun2024you,gloeckle2024betterfasterlarge}, prompt compression \cite{jiang-etal-2023-llmlingua,pan-etal-2024-llmlingua}, and KV cache optimization \cite{NEURIPS2023_6ceefa7b,yang-etal-2024-pyramidinfer,hooper2024kvquant10millioncontext}, among others. A specific approach within KV cache optimization, known as KV cache eviction, achieves efficiency improvements by exploiting characteristics and sparsity within attention mechanisms. Nevertheless, these methodologies predominantly concentrate on leveraging intrinsic sparsity of LLM itself, without considering external information to facilitate better optimization.

Previous study \cite{chen2021bert2bertreusablepretrainedlanguage} has indicated a significant similarity in attention patterns between small and large models within the BERT architecture. In Appendix \ref{sec:appendix_1}, we further substantiate that this similarity is also present in LLMs. Building upon this characteristic, this paper introduces an efficient inference technique through attention mapping between small language model (SLM) and the larger one. It is important to note that our proposed method achieves dual benefits of accelerated attention computation and reduced KV cache comsumption, and is orthogonal to most existing KV cache optimization methods. As depicted in Figure \ref{fig:reduce}, the utilization of $H_2O$ \cite{NEURIPS2023_6ceefa7b} facilitates KV cache compression at the token level, whereas our method can achieve further compression at the model layer level.

In this study, we investigate the methodology to achieve attention mapping between SLM and LLM while maintaining the performance of the original LLM. Initially, we evaluate the impact of various similarity metrics applied to attention matrices on language modeling efficacy. Following this, we explore how mapping at different layers of the LLM influences its language-understanding capabilities, thereby identifying the most appropriate layers for mapping. Finally, we demonstrate that the established mapping relation remains consistent throughout the inference process, enabling it to be constructed during the prefill stage and subsequently utilized in subsequent decode stage. This method of establishing mapping also facilitates dynamic adaptation to evolving contexts.

Based on these experiments and observations, we introduce the efficient \textbf{I}nference through \textbf{A}ttention \textbf{M}apping (\textbf{IAM}) framework. This framework effectively captures the similarities of attention patterns between SLM and LLM to dynamically establish mapping across varying contexts. Consequently, LLM can perform efficient inference without calculating portions of attention matrices, thereby achieving dual benefits of reduced GPU memory usage for KV cache and decreased computational requirements in attention mechanisms. We first comprehensively evaluate the performance preservation of IAM across four kinds of scenarios. Experimental results indicate that with a 30\% mapping ratio, the model maintains performance close to lossless, while at a 50\% mapping ratio, it also retains high capability levels. Efficiency evaluations across different inference scenarios demonstrate that IAM achieves an average reduction of 22.1\% in KV cache usage and an average acceleration of 11\% in inference speed. Our other experimental results indicate that the IAM is generalizable on other series of LLMs and compatible with existing KV cache optimization methods.
\section{Related Work}

Due to the widespread adoption of technologies such as RAG and the recent emergence of reasoning models like DeepSeek-R1, the demand for handling long contexts has significantly increased. The self-attention mechanism necessitates computing attention between the current token and every preceding token, leading to the common practice of storing previous tokens' KV states (KV cache) to avoid recomputation. However, this approach has become a primary bottleneck for managing long contexts. 

One category of methods focuses on efficiently storing and transmitting large amounts of KV cache with constrained hardware. For instance, tensor parallelism \cite{shoeybi2020megatronlmtrainingmultibillionparameter} distributes attention heads and pipeline parallelism \cite{10.5555/3454287.3454297} distributes attention layers across multiple GPUs, enabling horizontal scaling of KV cache storage capacity by adding more GPUs. When GPU HBM is insufficient, some techniques \cite{sheng2023flexgenhighthroughputgenerativeinference} concentrate on efficiently offloading the KV cache to CPU memory. Mooncake \cite{qin2024mooncakekvcachecentricdisaggregatedarchitecture} advances this concept further by proposing a multi-level caching strategy centered around the KV cache. At the CUDA optimization level, FlashAttention \cite{dao2023flashattention2fasterattentionbetter} reduces the number of read/write operations between GPU HBM and GPU cache, while PagedAttention \cite{kwon2023efficientmemorymanagementlarge} employs virtual memory management techniques to minimize memory fragmentation in GPU HBM. These approaches primarily aim to optimize hardware capabilities and KV cache requirements from a system perspective, without addressing KV cache reduction from an algorithmic perspective.

Another category of methods focuses on leveraging the inherent similarities and sparsity within the attention mechanism. Techniques such as KVQuant \cite{hooper2024kvquant10millioncontext} exploit redundancies in numerical representations to propose quantization of the KV cache, thereby reducing storage requirements and enhancing load speeds. StreamingLLM \cite{xiao2024efficientstreaminglanguagemodels} achieves closely unlimited input by reserving both the initial and the most recent KV cache of tokens. $H_2O$ \cite{NEURIPS2023_6ceefa7b} observed that the accumulated attention scores of all tokens follow a power-law distribution, indicating that only a small subset of tokens is highly significant in the generation. Scissorhands \cite{NEURIPS2023_a452a7c6} revealed the persistence of importance, indicating that tokens identified as important in initial remain significant throughout subsequent stages of inference. PyramidInfer \cite{yang-etal-2024-pyramidinfer} further explores the distinct attention characteristics across different layers within LLM, and identified that deeper layers exhibit greater redundancy. These findings help design effective KV cache eviction approaches. Meanwhile, KVMerger \cite{wang2024modeltellsmergeadaptive} leverages the high similarity of KV states observed across different datasets to design a Gaussian kernel weighted merging algorithm for merging KV Cache, thereby minimizing the loss introduced by dropping methods. However, these approaches primarily exploit the inherent similarities and sparsity within the model's attention mechanism and do not incorporate external information to enhance KV cache optimization, as IAM does. In addition, IAM is orthogonal to most of the aforementioned methods, due to its innovative approach of utilizing entire attention matrices of SLM and avoiding computations of attention mechanism of mapped layers in LLM.

There are also some works about speculative decoding \cite{xia2023speculativedecodingexploitingspeculative,cai2024medusasimplellminference}, which employs a SLM as a draft model, followed by validation from a LLM to determine whether to adopt its predictions. Similar to our method, this approach aims to accelerate inference through the collaboration of small and large models. However, the core mechanisms differ: speculative decoding focuses on aligning the next-word prediction probability distributions between the SLM and LLM, while IAM relies on the high similarity of attention matrices between the SLM and LLM. Additionally, the LLM in speculative decoding can also benefit from the IAM method to enhance performance, achieving a synergistic effect.
\section{Methodology}

\iffalse
IAM centers on the mapping of attention matrix based on similarity between language models with different scales, thereby reducing resource consumption and facilitating efficient inference. However, performing this mapping while maintaining or closely approximating the original model's performance presents a significant challenge. 
\fi

In this section, we begin by conducting a series of experiments aimed at providing valuable insights into attention mapping. Building on these findings, we then introduce the IAM framework.

\subsection{How to Measure Similarity}
\label{How to Measure Similarity}
A proper metric of similarity is crucial in attention mapping. It helps to capture the significant pattern within the attention matrix, thus minimizing the bias introduced by mapping in the forward process of LLM. For vectors $\mathbf{x}$ and $\mathbf{y}$, formed by flattening the lower triangular part of attention matrices of SLM and LLM, we consider the following similarity measure:

\noindent\textbf{Cosine Similarity.} The cosine similarity is defined as:

\begin{equation}
cos<\mathbf{x},\mathbf{y}> = \frac{\mathbf{x} \cdot \mathbf{y}}{\|\mathbf{x}\| \|\mathbf{y}\|}
\end{equation}

\noindent\textbf{Minkowski Distance.} The Minkowski Distance is defined as:

\begin{equation}
d(\mathbf{x}, \mathbf{y}) = \left( \sum_{i=1}^{n} |x_i - y_i|^p \right)^{\frac{1}{p}}
\end{equation}

where $x_i$ and $y_i$ is the element in vector $\mathbf{x}$ and $\mathbf{y}$, and $p$ is the hyperparameter.

\noindent\textbf{Pearson correlation.} The Pearson correlation is a common measure of linear correlation between two variables. It is defined as:
\begin{equation}
r_{\mathbf{xy}} = \frac{\sum (x_i - \bar{x})(y_i - \bar{y})}{\sqrt{\sum (x_i - \bar{x})^2} \sqrt{\sum (y_i - \bar{y})^2}}
\end{equation}
where $x_i$ and $y_i$ is the element of vector $\mathbf{x}$ and $\mathbf{y}$, and $\bar{x}$ and $\bar{y}$ represents mean of $\mathbf{x}$ and $\mathbf{y}$.

We utilize Qwen2-72B as the target LLM and consider two SLMs with different scales, Qwen2-0.5B and Qwen2-7B, as sources of attention matrices for mapping. We measure the log perplexity of the LLM on WikiText-v2 \cite{merity2016pointersentinelmixturemodels} after performing attention mapping using different metrics of similarity. The experimental results are presented in Table \ref{tab:Metric}. It is noted that attention mapping is applied to the layers in the latter half of the LLM.

From the experimental results, it can be observed that cosine similarity achieves the best performance for both mappings from Qwen2-0.5B to Qwen2-72B and from Qwen2-7B to Qwen2-72B. Minkowski Distance with p = 1 also performs nearly as well as the best metric. Pearson correlation is proven unsuitable as the similarity metric due to its significantly higher perplexity. We also explore compensating the differences of the L2 norm after selecting the mapping matrix based on cosine similarity, which is named "Cosine with Norm" in Table \ref{tab:Metric}. However, the experimental results indicate that this approach does not consistently and significantly enhance the performance of cosine similarity. Moreover, introducing the variable of L2 norm adds complexity to the mapping in the inference process and is unfavorable for the generalization across different contexts. Therefore, we finally choose cosine similarity as the similarity metric.

\begin{table}[]
\centering
\resizebox{1\columnwidth}{!}{
\begin{tabular}{clc}
\toprule
\multicolumn{1}{l}{Mapping} & Metric                      & Perplexity (log)\\ \midrule
\multirow{4}{*}{\begin{tabular}[c]{@{}c@{}}Qwen2-0.5B\\ to\\ Qwen2-72B\end{tabular}} & Cosine & 2.577 \\
                            & Pearson         & 4.976      \\
                            & Minkowski (p = 1)
                            & 2.589 \\
                            & Minkowski (p = 2)                    & 2.878      \\
                            & Cosine with Norm & \textbf{2.562}      \\ \midrule
                            
\multirow{4}{*}{\begin{tabular}[c]{@{}c@{}}Qwen2-7B\\ to\\ Qwen2-72B\end{tabular}}   & Cosine & \textbf{2.414} \\
                            & Pearson         & 5.376      \\
                            & Minkowski (p = 1)
                            & 2.427 \\
                            & Minkowski (p = 2)                     & 2.514      \\
                            & Cosine with Norm & 2.421      \\ \midrule
\multicolumn{2}{c}{Qwen2-72B (Original)}                             & 2.136      \\ \bottomrule
\end{tabular}}
\caption{The log perplexity values after conducting attention mapping using different similarity metrics. The original log perplexity is also provided in the bottom for comparison.}
\label{tab:Metric}
\end{table}

\subsection{Which Layers to Map}
\label{sec:layer_select}
In this section, we explore the methodology for selecting suitable layers to map, with the objective of retaining model performance. For our purposes, we utilize Qwen2-72B as the target LLM for mapping which comprises 80 transformer layers. We partitioned these layers into 10 sub-blocks (8 layers each) and make them mapped sequentially, followed by an evaluation of its performance using the MMLU benchmark. 

\begin{figure}[!htb]
\centering
\includegraphics[scale=0.24]{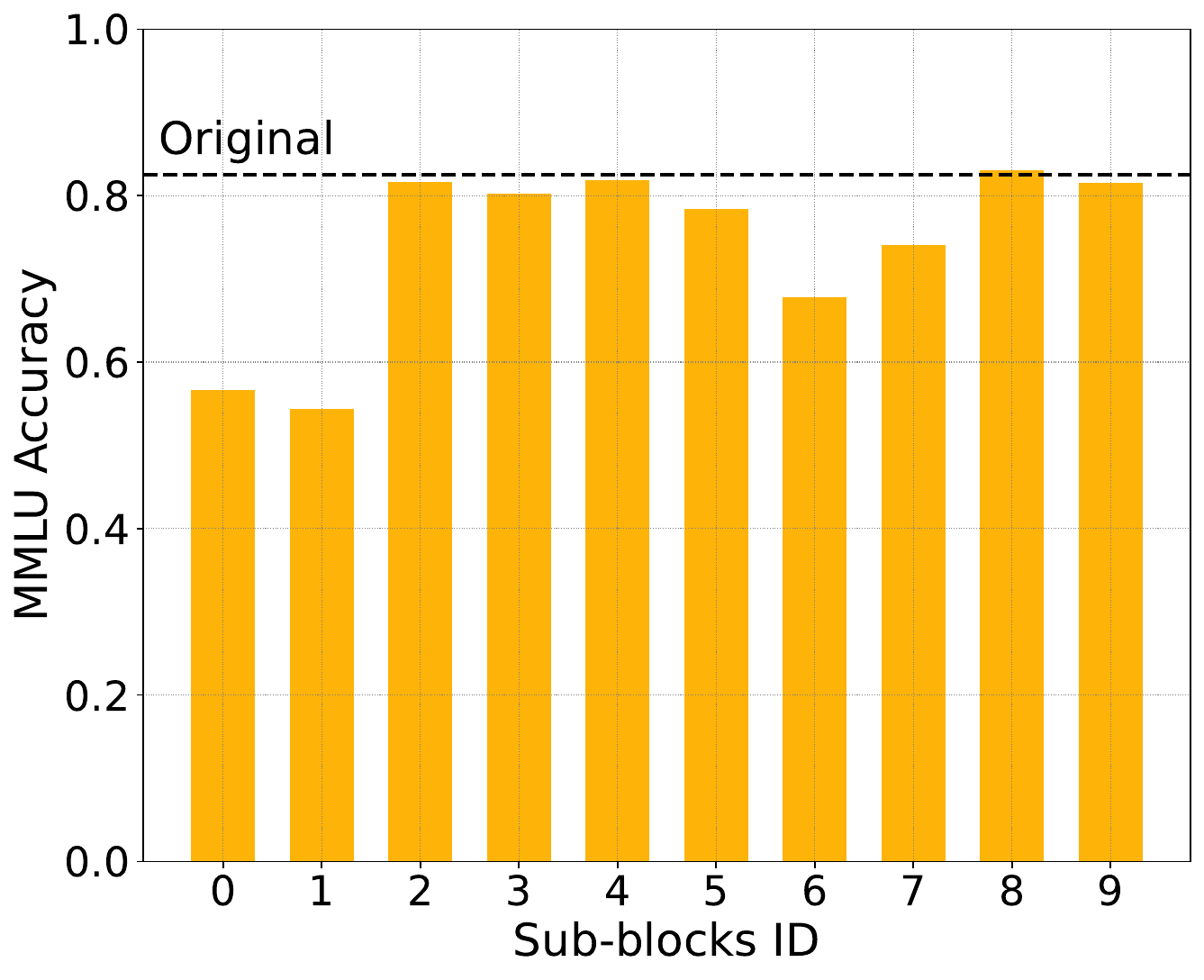}
\caption{The performance on MMLU benchmark after mapping different attention layers of Qwen2-72B.}
\label{fig:layer_select_mmlu}
\end{figure}

The experimental results are shown in Figure \ref{fig:layer_select_mmlu}. The experimental results indicate the presence of two optimal mapping regions within the Qwen2 model architecture: one situated in the final two sub-blocks (layers 64 - 80) and another located in No.2 to No.4 sub-blocks  (layers 16 - 40). These regions maintain performance levels comparable to those of the original model after attention mapping. Conversely, mapping from the beginning (layers 0 - 16) is entirely impractical due to the significant performance degradation it induces. 

We also examine the effectiveness of the mapping strategy on other tasks, using linear correlation to verify whether a well-performing strategy derived from one task (e.g., MMLU) can generalize to others. The experimental results are shown in Table 3. Noting that lower log perplexity indicates better performance, we take the reciprocal of this metric to align its trend with that of other tasks for consistency. As can be seen, the results of mapping different sub-blocks on the MMLU task show strong similarity with those on other tasks, as evidenced by the high Pearson correlation coefficients between them.

\begin{table*}[]
\resizebox{2.1\columnwidth}{!}{
\begin{tabular}{@{}lccccccccccc@{}}
\toprule
\textbf{Benchmark} &
  \textbf{Block 0} &
  \textbf{Block 1} &
  \textbf{Block 2} &
  \textbf{Block 3} &
  \textbf{Block 4} &
  \textbf{Block 5} &
  \textbf{Block 6} &
  \textbf{Block 7} &
  \textbf{Block 8} &
  \textbf{Block 9} &
  \textbf{Pearson} \\ \midrule
MMLU               & 0.567 & 0.544 & 0.816 & 0.802 & 0.819 & 0.784 & 0.678 & 0.741 & 0.83  & 0.815 & -     \\
1/Perplexity (log) & 0.218 & 0.288 & 0.326 & 0.350 & 0.364 & 0.374 & 0.383 & 0.388 & 0.395 & 0.403 & 0.741 \\
HotpotQA (Norm.)   & 0.386 & 0.415 & 0.924 & 1.278 & 1.311 & 0.560 & 0.779 & 1.108 & 0.907 & 1.032 & 0.763 \\
GovReport (Norm.)  & 0.160 & 0.259 & 0.797 & 0.996 & 1.007 & 1.090 & 0.639 & 0.806 & 1.014 & 1.047 & 0.950 \\ \bottomrule
\end{tabular}}
\caption{The performance on different benchmarks after
mapping different attention layers of Qwen2-72B. The high Pearson correlation coefficients indicate the consistency of the impact on different tasks with the mapping strategy.}
\label{tab:map_str}
\end{table*}

Based on these observations, IAM adopts the following mapping strategy for Qwen2-72B: according to the user-specified mapping ratio, mappings are first performed sequentially from the last layer of the model backward to layer 64. If the specified mapping ratio is still not achieved, the mapping continues from layer 16 onward, proceeding backwards until the requirement is met. This approach ensures a balance between maintaining model performance and achieving the desired mapping efficiency, leveraging the identified optimal regions for layer mapping.

\subsection{Consistency of Mapping}
\label{sec:consistency}
IAM establishes a mapping relationship between the SLM and the LLM based on similarity during the prefill stage, which is then utilized in the subsequent decode stage. Therefore, an essential guarantee is that the mappings established during the prefill stage remain consistent, which means mapping relationship should not significantly change as the inference progresses. To evaluate the consistency, we observe the proportion of times that the mapping does not change as consistency rate during the autoregressive generation. We also use the WikiText-v2 datasets as the prompt and set the max output tokens as 500.

%and more statistical information can be referred to Appendix. 

\begin{figure}[!htb]
\centering
\includegraphics[scale=0.24]{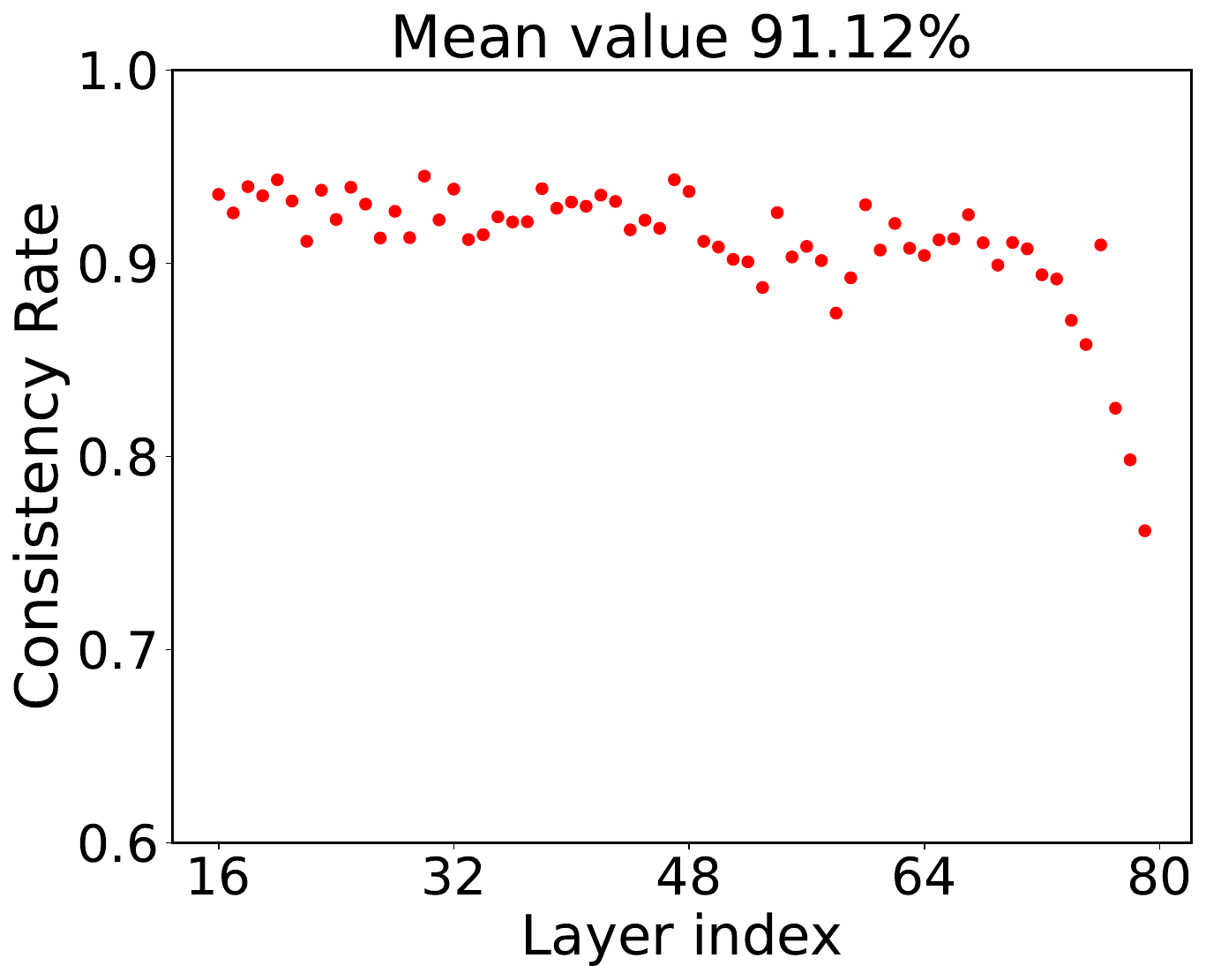}
\caption{The average consistency rate from the 16th layer onward. A higher consistency rate indicates that the mapping established in prefill stage remains more stable and undergoes fewer changes during decoding.}
\label{fig:consistency}
\end{figure}

The experimental results are shown in Figure \ref{fig:consistency}, where we averaged the consistency rate within layers. It can be seen that a high level of consistency rate is maintained across layers. This indicates that the mapping established during the prefill stage remains stable and reliable throughout the decode stage, ensuring relatively small imprecision of mapping during the dynamic generation.

\begin{figure*}[!htb]
\centering
\includegraphics[scale=0.48]{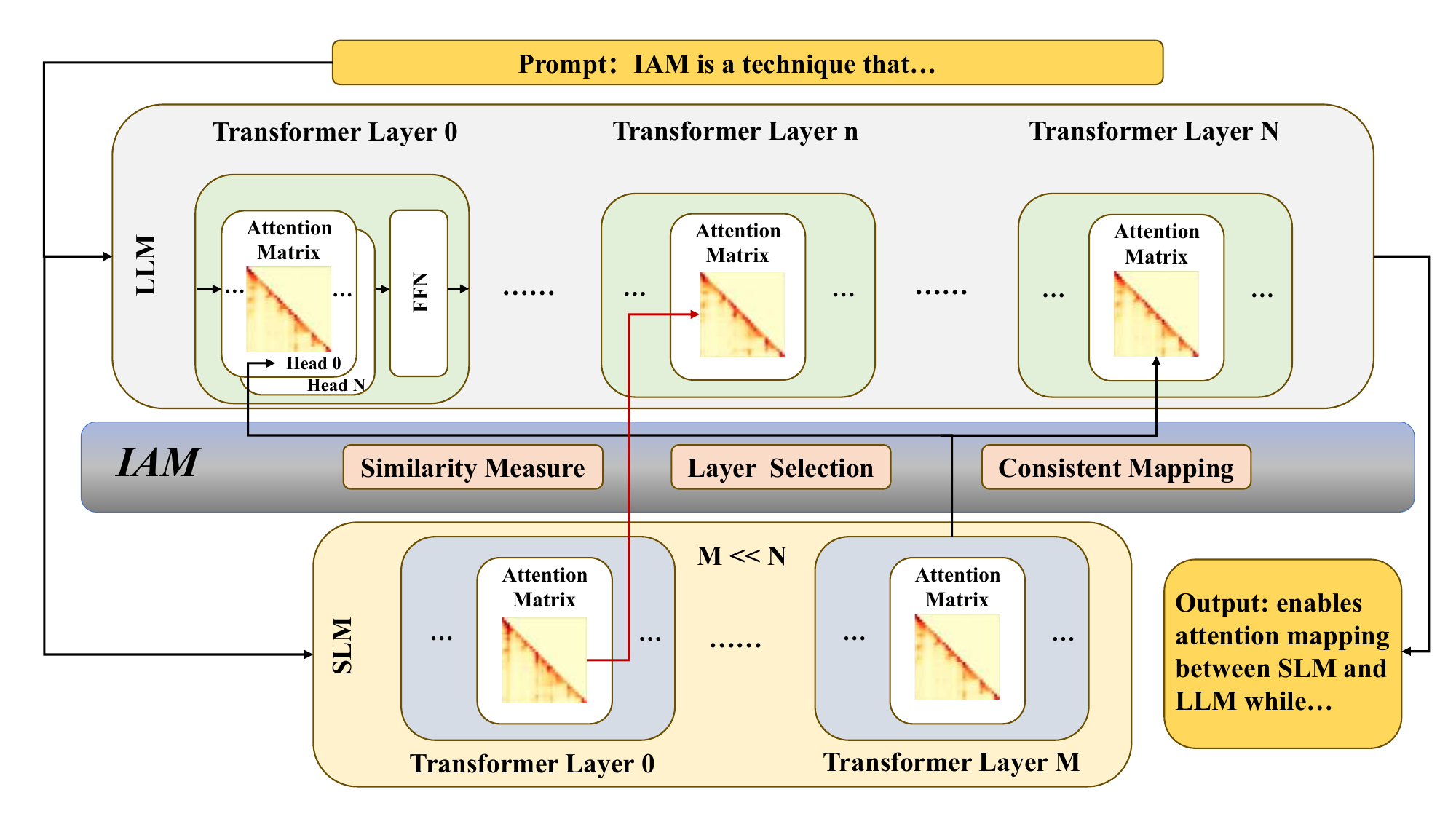}
\caption{Framework of IAM.}
\label{fig:framework}
\end{figure*}

\subsection{IAM Framework}

Based on the preceding experiments and observations, we present the framework of IAM in this section, as shown in Figure \ref{fig:framework}. IAM initially establishes mappings during the prefill stage. Given a LLM $M$ and a SLM $M_s$, the prompt is first passed through both models to obtain the attention matrices respectively. Specifically, the attention matrices of $M$ are denoted as \( A_{i} \in \mathbb{R}^{N \times N} \), \( 0 < i < L \cdot D \), and those in $M_s$ are denoted as \( A'_{j} \in \mathbb{R}^{N \times N} \), \( 0 < j < l \cdot d \). \( L \) and \( l \) represent the number of layers in the $M$ and $M_s$. \( D \) and \( d \) denote the number of attention heads per layer in each model. \( N \) represents the number of tokens in the prompt. Due to differences in model architectures, we have $l \cdot d < L \cdot D$. And then, pairwise similarity between the attention matrices from $M$ and $M_s$ is computed by:

\begin{equation}
S(A_i, A'_j) = \frac{\text{Tr}(A_i^T A'_j)}{\|A_i\|_F \|A'_j\|_F}
\label{eq:simular}
\end{equation}

\(\text{Tr}(\cdot)\) denotes the trace of the matrix and \(\|\cdot\|_F\) denotes the Frobenius norm. Finally, the mapping function is obtained by:

\begin{equation}
f(i) = \arg\max_j S(A_i, A'_j)
\label{eq:mapping}
\end{equation}

In the decode stage, IAM utilizes the established mapping $f(i)$ to perform the attention mapping, which can be represented as:

\begin{equation}
A_i \leftarrow A'_{f(i)}
\label{eq:replace}
\end{equation}

Finally, the forward process of the model based on IAM can be represented as:
\begin{equation}
\mathbf{Y} = M(\mathbf{X}; \theta^{M}, A_i \leftarrow A'_{f(i)})
\end{equation}
where $\mathbf{X}$ and $\mathbf{Y}$ denotes the input and output respectively.

After establishing the IAM framework, we perform instruction tuning to mitigate the performance degradation caused by differences in the attention matrices. Specifically, we utilized the Alpaca dataset \cite{alpaca} for this purpose. More details about training can be found in Appendix \ref{app:train}. The optimization objective for $M_s$ can be formulated as:

$$
\min_{\theta^{M_s}} \mathbb{E}_{(\mathbf{X}, \mathbf{T}) \sim \mathcal{D}} \left[ \mathcal{L}(M(\mathbf{X}; \theta^{M_s}, A_i \leftarrow A'_{f(i)}), \mathbf{T}) \right]    
$$

It is also noted that to prevent instability in the mapping established during the prefill stage due to overly short prompt, we employ a delayed establishment mechanism. This means that mapping is only initiated and performed once the number of preceding tokens 
$N$ exceeds a specified threshold. Similarly, for scenarios involving long contexts, based on observations of consistency in Section \ref{sec:consistency}, we truncate the sequences when calculating similarity to avoid inaccuracies that arise from high-dimensional data. The detailed procedure of IAM can be referred to Algorithm \ref{alg:IAM}.

\begin{algorithm}[t]
	\caption{IAM} 
	\textbf{Input}: A target large language model $M$; A small language model $M_s$; Prompt ${\bm{x}}$; Establishing threshold $\tau_e$; Truncation threshold $\tau_t$; Mapping ratio $R$; Max output lenth $L_{max}$.
	\begin{algorithmic}[1]

		\State Calculate token lenth of prompt as $len(\bm{x})$
		\If {$len(\bm{x}) > \tau_t$}
			\State Truncate lenth of prompt to $\tau_t$
		\ElsIf {$len(\bm{x}) < \tau_e$}
			\State Autoregressive generation ${\bm{y}} = \{{y}_i\}_{i=1}^{{k}}$ by $M$ until $len(\bm{x}) + k >= \tau_e$
		\EndIf
		\State Forward process by $M(\bm{x})$ and $M_s(\bm{x})$ and get attention matrices respectively
		\State Calculate pairwise similarity via Eq.(\ref{eq:simular})
		\State Select mapping layers according to $R$
		\State Establish mapping via Eq.(\ref{eq:mapping}) 
		\While{${y}_i \neq eos$ and $L < L_{max}$}
			\State Replace attention matrix of $M$ via Eq.(\ref{eq:replace}) 
			\State Autoregressive generation ${\bm{y}} = \{{y}_i\}_{i=k}^{{L}}$ by $M(\bm{x},\bm{y})$
		\EndWhile	
\end{algorithmic}
\textbf{Output}: Output token sequences ${\bm{y}} = \{{y}_i\}_{i=1}^{{L}}$
\label{alg:IAM}
\end{algorithm}

\iffalse
\begin{algorithm}[t]
	\caption{IAM} 
	\textbf{Input}: A target large language model $M$; A small language model $M_s$; Prompt ${\bm{x}}$; Establishing threshold $\tau_e$; Truncation threshold $\tau_t$; Mapping ratio $R$; Max output lenth $L_{max}$.
	\begin{algorithmic}[1]
		\State Forward process by $M(\bm{x})$ and $M_s(\bm{x})$ and get attention matrices respectively
        \If {$len(\bm{x}) > \tau_t$}
		\State Truncate attention matrices to $\tau_t \times \tau_t$
		\ElsIf {$len(\bm{x}) < \tau_e$}
		\State Autoregressive generation ${\bm{y}} = \{{y}_i\}_{i=1}^{{k}}$ by $M$ until $len(\bm{x}) + k >= $\tau_e$
		\EndIf
        
		\State Calculate pairwise similarity via Eq.(\ref{eq:simular})
		\State Select mapping layers according to $R$
		\State Establish mapping via Eq.(\ref{eq:mapping}) 
		\While{${y}_i \neq eos$ and $L < L_{max}$}
			\State Replace attention matrix of $M$ via Eq.(\ref{eq:replace}) 
			\State Autoregressive generation ${\bm{y}} = \{{y}_i\}_{i=k}^{{L}}$ by $M(\bm{x},\bm{y})$
		\EndWhile	
\end{algorithmic}
\textbf{Output}: Output token sequences ${\bm{y}} = \{{y}_i\}_{i=1}^{{L}}$
\label{alg:IAM}
\end{algorithm}

\fi
\section{Experiments}
\subsection{Experiments Setup}

\paragraph{Datasets}We comprehensively evaluate the utility preservation of IAM from four kinds of scenarios: 1) Language modeling: we measure the perplexity on WikiText-v2 \cite{merity2016pointersentinelmixturemodels}. 2) Language understanding: We evaluate performance on the MMLU \cite{hendrycks2021measuringmassivemultitasklanguage}. 3) QA task: we assess QA performance using the HotpotQA dataset \cite{yang2018hotpotqadatasetdiverseexplainable} and adopt F1 score as evaluation metric. 4) Long context: we utilize GovReport benchmark \cite{huang-etal-2021-efficient} to test the long text summarization capacity.

\paragraph{Implementation Details}Our experiments are conducted on two types of LLMs. The main results are obtained using the models of the Qwen2 series\footnote{https://github.com/QwenLM/Qwen}. Specifically, we use Qwen2-72B as target LLM and two SLMs with different scales: Qwen2-0.5B and Qwen2-7B. We also test the LLaMA3 series models\footnote{https://github.com/meta-llama/llama3} to study the IAM's adaptation to different series LLMs in section \ref{sec:llama}. For these experiments, the SLM is LLaMA 3.2-1B, and the LLM is LLaMA 3.1-70B. 

All experiments are performed on 8 NVIDIA A100 (80GB) GPUs. We use greedy decoding to ensure the stability of experimental results. For generative tasks, we limit the maximum output tokens to 512 and set the repetition penalty as 1.2. We also set the establishing threshold $\tau_e$ and the truncation threshold $\tau_t$ in IAM algorithm as 20 and 100 respectively. The other experimental environment includes the following configurations: CUDA version 12.0, PyTorch version 2.4.0, and HuggingFace's Transformers\footnote{https://github.com/huggingface/transformers} with version 4.45.1.

\subsection{Benchmark Results}

\begin{figure*}[t]
\centering
\subfloat{\includegraphics[scale=0.14]{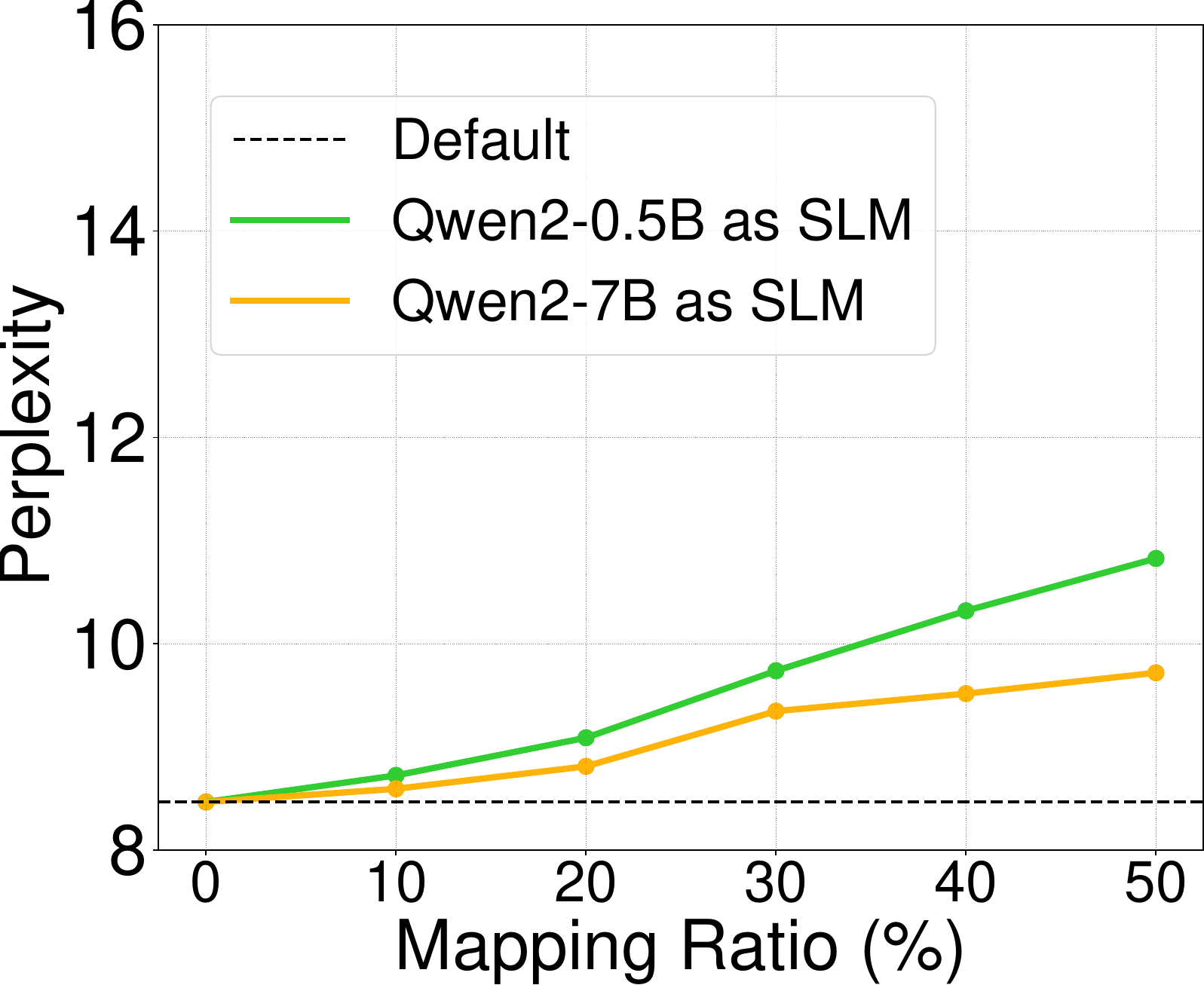}}
\hspace{2pt}
\subfloat{\includegraphics[scale=0.14]{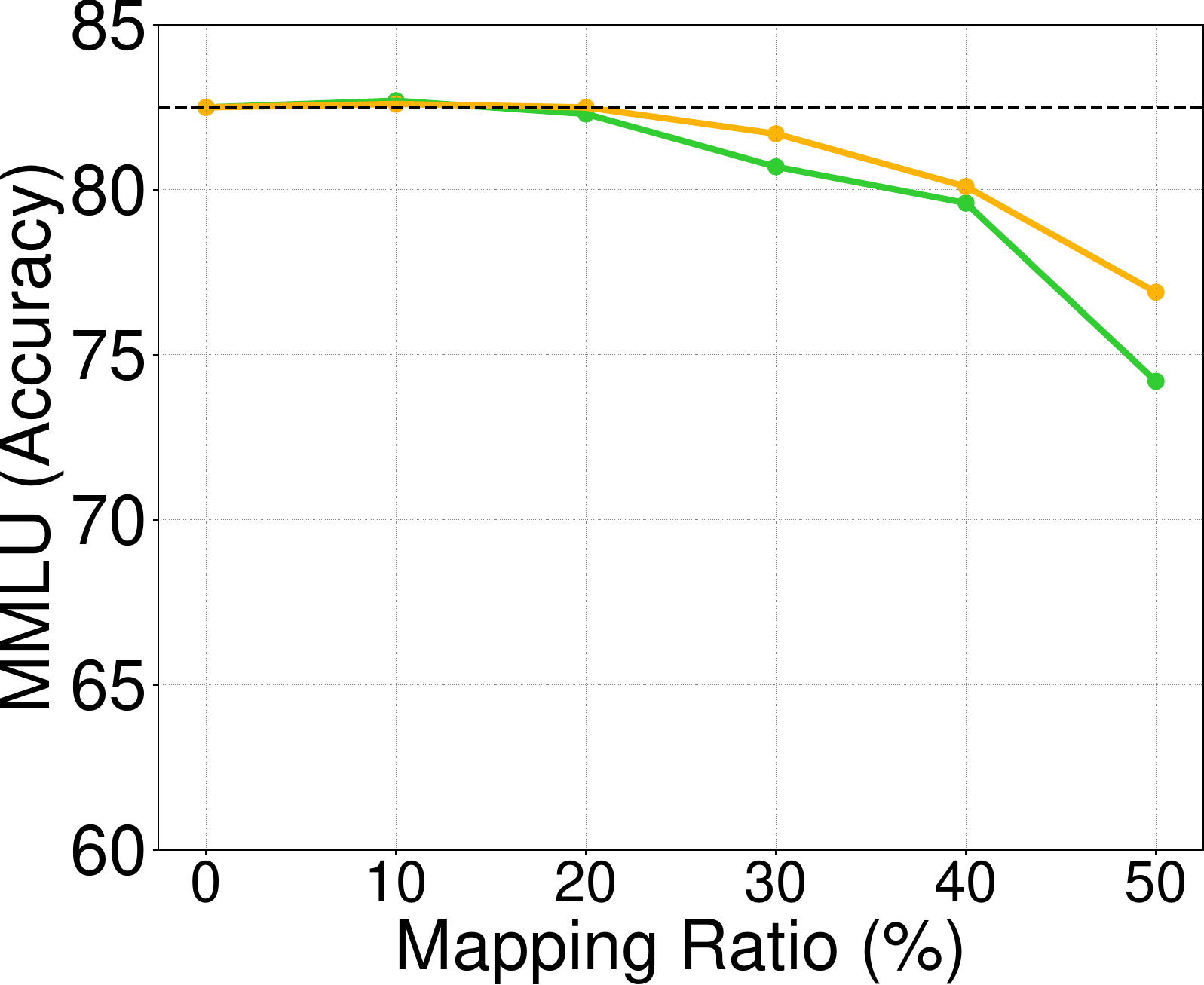}}
\hspace{2pt}
\subfloat{\includegraphics[scale=0.14]{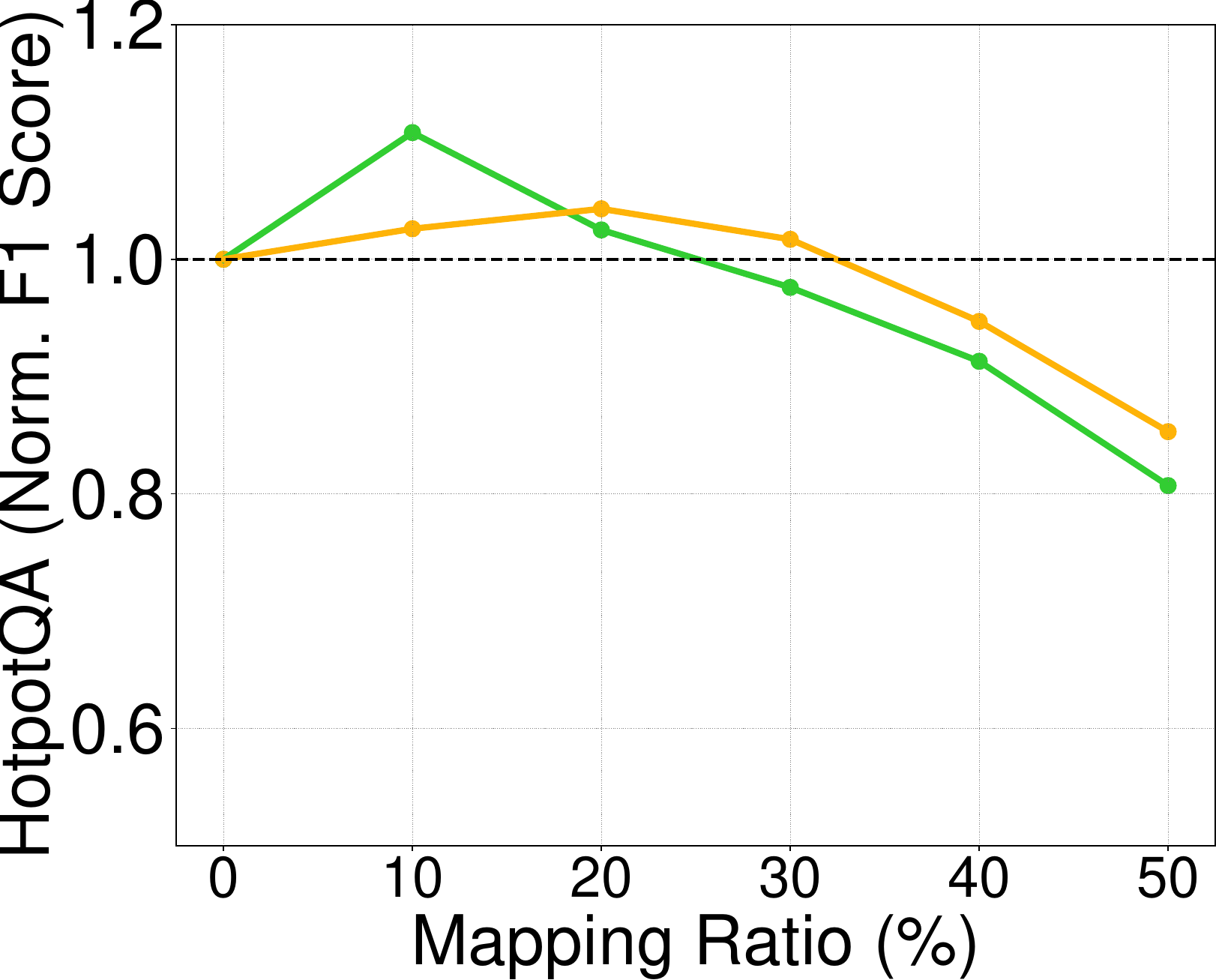}}
\hspace{2pt}
\subfloat{\includegraphics[scale=0.14]{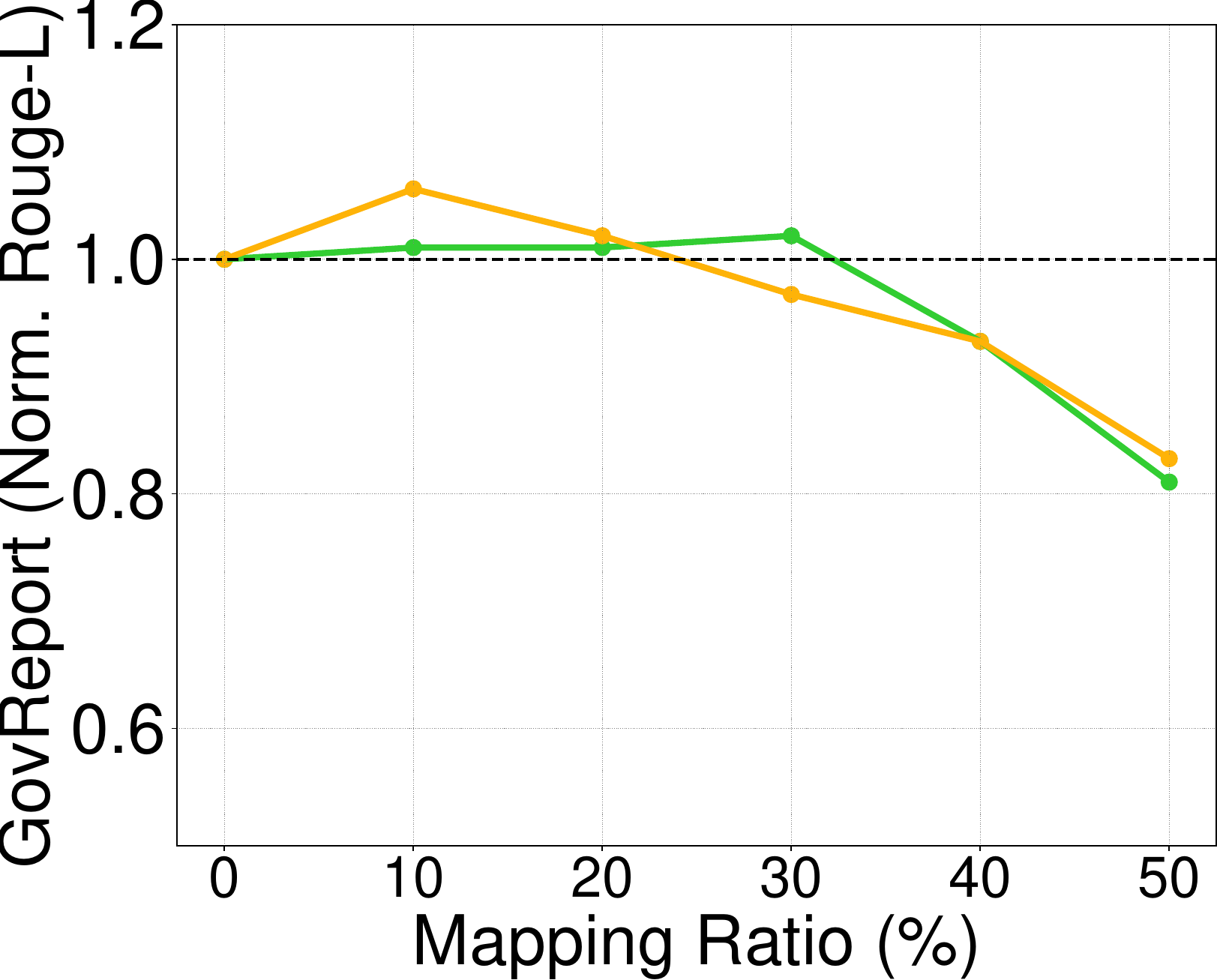}}
\caption{Benchmark results of IAM with mapping ratio varying from 0 to 50\%. Default represents using the original LLM without mapping.}
\label{fig:main}
\end{figure*}

In Figure \ref{fig:main}, we evaluate the performance of IAM in four kinds of scenarios with mapping ratio varying from 0 to 50\%. It can be seen that IAM maintains high language understanding and generation quality with much less GPU memory consumption. Specifically, at a mapping ratio of 30\%, IAM achieves almost the same performance as the original model. At a mapping ratio of 50\%, it still retains high capability levels. 

It is also important to note that using Qwen2-7B as the SLM yields better performance. We conjecture that this is because using the bigger model with more source attention matrices can improve maximum similarity between mapping layers, which will reduce the loss of mapping. This implies that there is a trade-off between efficiency and model performance.

\subsection{Efficiency Analysis}

To evaluate the efficiency of IAM, we consider two common scenarios: 1) Multi-user Concurrency: In this scenario, the context length is set to a prefill length of 512 tokens and a maximum generation length of 512 tokens, with a batch size of 64. 2) Long Context: In this scenario, the context length is set to a prefill length of 8192 tokens and a maximum generation length of 512 tokens, with a batch size of 8. In each scenario, we analyze both memory efficiency and compute efficiency, and thus derive a detailed view of the impact of IAM on end-to-end performance. The LLM used in this part of the experiment is Qwen2-72B while the SLM is Qwen2-0.5B, and the mapping ratio is set to 0.5.

\begin{table*}[]
\centering
\caption{Evaluation on efficiency of IAM. \textbf{Bsz}: batch size. \textbf{Lenth}: prefill length + decode lenth. \textbf{KV Mem.}: GPU memory usage (GB) of the KV cache. \textbf{TPOT}: average time (s) per output token in decode stage. \textbf{TTFT}: time (s) to first token. \textbf{Thr.}: End-to-end throughput (token/s)}
\begin{tabular}{c|cc|cccc}
\toprule
\textbf{Method}        & \textbf{Bsz}                 & \textbf{Lenth}                    & \textbf{KV Mem}.                 & \textbf{TPOT}         & \textbf{TTFT}         & \textbf{Thr.}                  \\ \midrule
Default       & \multirow{2}{*}{64} & \multirow{2}{*}{512+512}  & 158.7 (100\%)           & 0.196 (1.0x)  & 2.67 (1.0x)  & 636 (1.0x)           \\
IAM &                     &                           & \textbf{124.2 (78.3\%)} & 0.176 (1.11x) & 2.39 (1.12x) & \textbf{708 (1.11x)} \\ \midrule
Default       & \multirow{2}{*}{4}  & \multirow{2}{*}{8192+512} & 187.4 (100\%)           & 0.216 (1.0x)  & 6.72 (1.0x)  & 297 (1.0x)           \\
IAM &                     &                           & \textbf{143.1 (77.5\%)} & 0.197 (1.10x) & 5.74 (1.17x) & \textbf{326 (1.10x)} \\ \bottomrule
\end{tabular}
\label{tab:eff}
\end{table*}

\paragraph{Memory Efficiency}

With IAM, the mapped layers in the LLM do not need to store the K cache because they do not compute attention matrices. This results in a reduction of memory usage. Although the introduction of SLM with its parameters and kv cache also incurs additional overhead. It is acceptable since the SLM’s memory is significantly smaller than that of the LLM. Referring to Table \ref{tab:eff}, even when accounting for the KV cache of the SLM, IAM achieves a 21.7\% reduction in KV cache usage in the multi-user concurrency scenario and a 22.5\% reduction in the long context scenario, which indicates the memory efficiency of IAM.

\paragraph{Compute Efficiency}

Similarly, in IAM, the mapped layers of LLM do not need to perform Q projection, K projection, and the multiplication of the QK matrix with softmax. These computations are particularly intensive during the prefill stage and often become a performance bottleneck. By using IAM, we can significantly reduce this computational load. Although introducing the SLM adds some computational overhead, this additional workload is significantly lower compared to the computations avoided in the LLM. As a result, the overall system still sees substantial benefits: In the multi-user concurrency scenario, the reduction in computational load leads to a 12\% decrease in TTFT. In the long context scenario, where the computational complexity of the QK matrix multiplication increases quadratically with sequence length, IAM provides even greater computational savings, resulting in a more significant 17\% reduction in TTFT.

\paragraph{End-to-end Efficiency}

The resulting memory and compute savings contribute to improved overall system efficiency. Benefit from the substantial reduction in K cache usage, the LLM experiences notable improvements in TPOT during the decode stage, with decreases of 11\% and 10\% in the multi-user concurrency and long context scenarios, respectively. Additionally, the significant reduction in computational load leads to improved TTFT during the prefill stage, with decreases of 12\% and 17\% in the same scenarios. These enhancements translate into improved end-to-end throughput performance. Compared to the default method, IAM achieves throughput improvements of 11\% and 10\% in the multi-user concurrency and long context scenarios, making it both practical and beneficial for various deployment scenarios.

\subsection{Different Series Models}
\label{sec:llama}

\begin{figure}[!htb]
\centering
\includegraphics[scale=0.24]{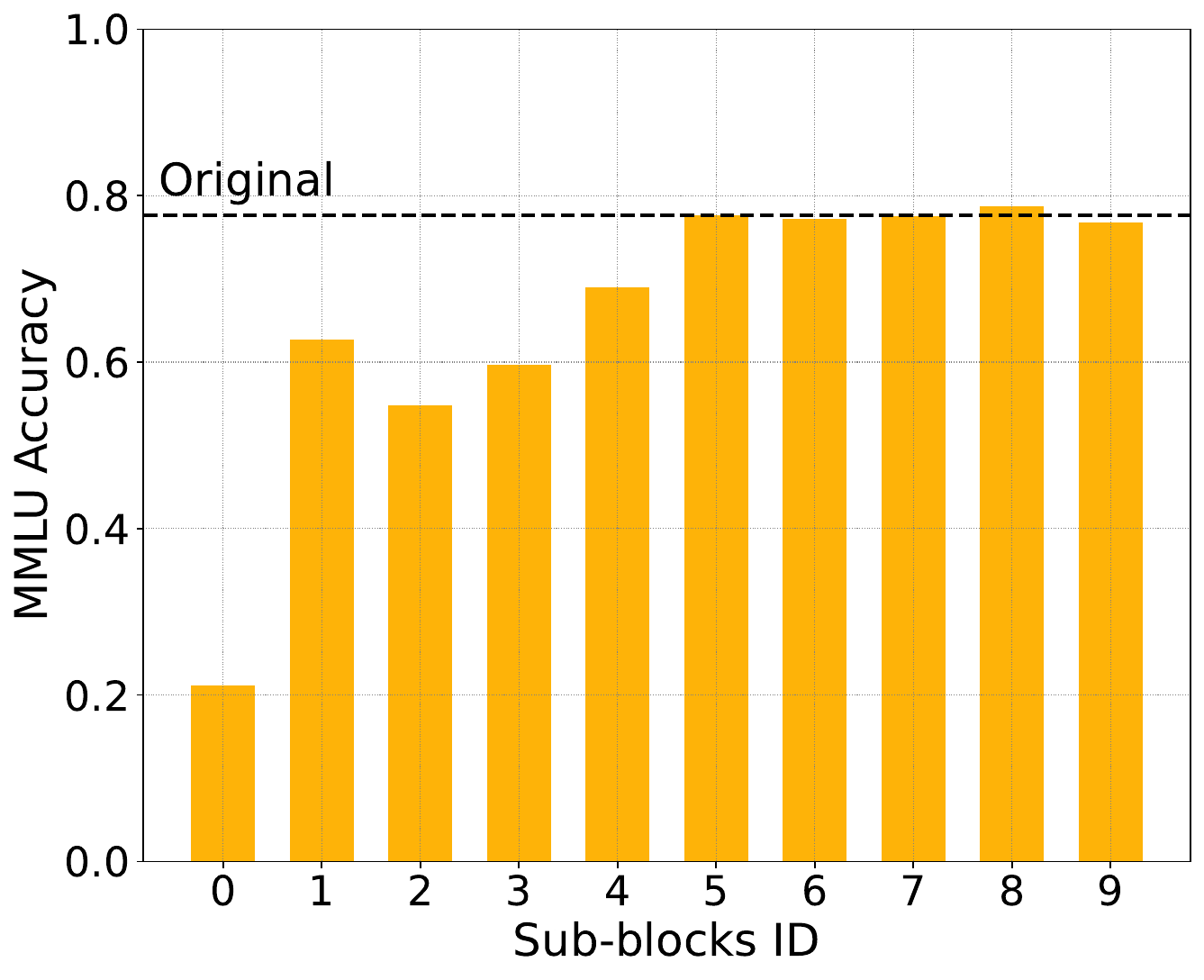}
\caption{The performance on MMLU benchmark after mapping different attention layers of LLaMA 3.1-70B.}
\label{fig:layer_select_llama}
\end{figure}

To evaluate the generalizability of the IAM, we conduct experiments on different series of models. In this experiment, the LLM is LLaMA 3.1-70B and the SLM is LLaMA 3.2-1B. The procedure of IAM for different series of models is largely followed the procedures detailed in the methodology section, with the primary modification being the re-identification of suitable layers within LLaMA 3.1-70B for mapping. Following the same approach described in Section \ref{sec:layer_select}, we test the suitable layers within LLaMA 3.1-70B and show results in Figure \ref{fig:layer_select_llama}. An interesting observation is that, unlike the Qwen2 series models which exhibit two optimal mapping regions, the LLaMA series models only have a single optimal mapping region located at the end of the model. Consequently, we adjusted our mapping strategy for the LLaMA series models: as the mapping ratio increases, mappings are sequentially established from the last layer backward toward earlier layers.

After selecting the appropriate layers for mapping in the LLaMA series models, we also evaluated their performance on the MMLU benchmark under various mapping ratios. The experimental results are shown in Figure 7. Even better than the performance with the Qwen2 series, IAM demonstrates excellent language comprehension capabilities on the LLaMA series models. 

\begin{figure}[!htb]
\centering
\includegraphics[scale=0.2]{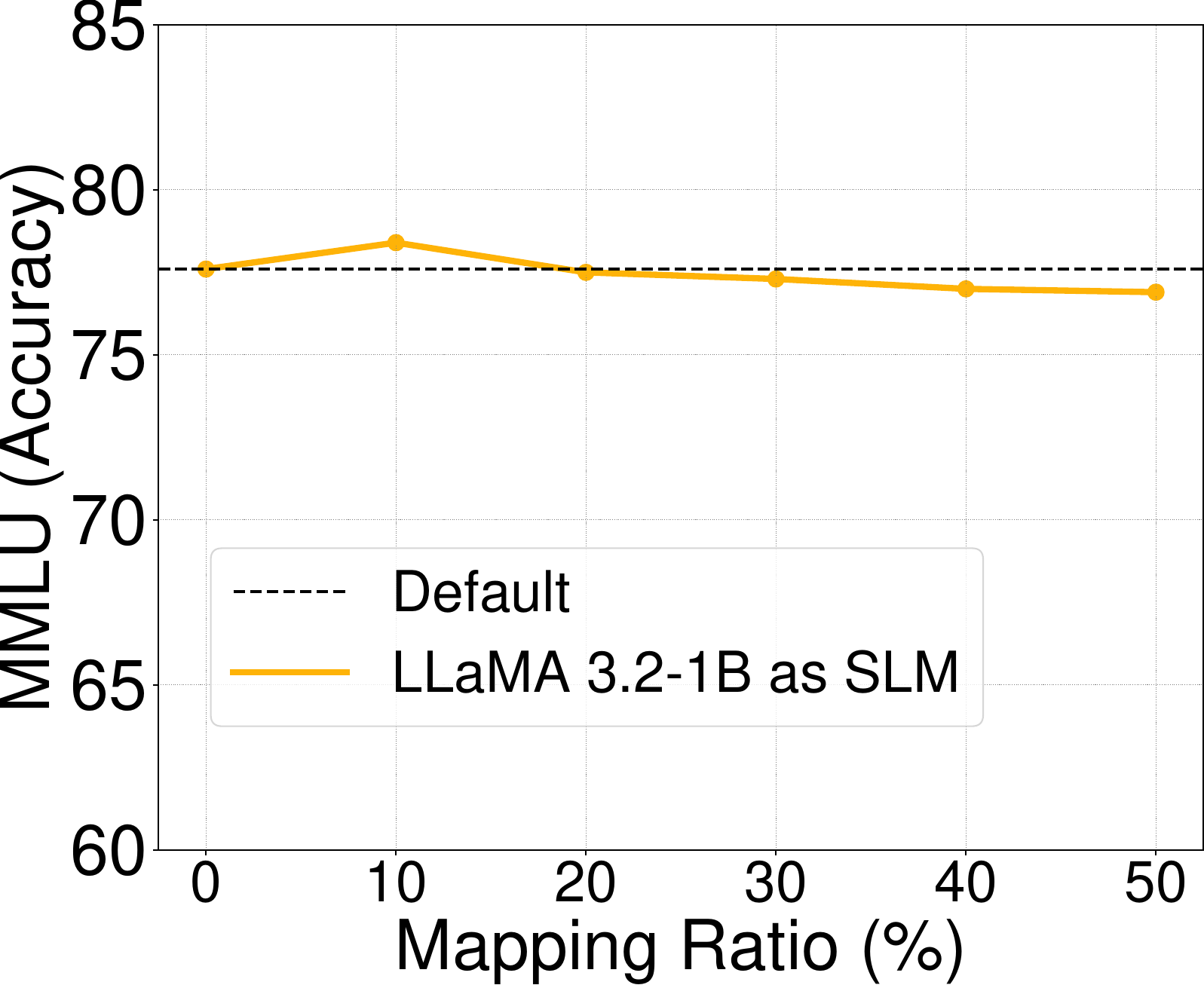}
\caption{The MMLU benchmark results of IAM on LLaMA series models.}
\label{fig:layer_select_llama}
\end{figure}

\subsection{Compatible with KV cache compression}

IAM is orthogonal to most existing KV cache compression methods due to its innovative approach of utilizing entire attention matrices of SLM for mapping, whereas existing methods mainly focus on token-level attention importance. To illustrate this, we take $H_2O$ \cite{NEURIPS2023_6ceefa7b} as an example. In Figure \ref{fig:reduce}, we demonstrate that, after the KV cache optimization of $H_2O$, IAM can further reduce KV cache consumption. We also evaluate the impact of using $H_2O$ on IAM's performance in terms of perplexity on the WikiText-v2 dataset, as shown in Figure 3. The KV cache budget of $H_2O$ is set to 80\%. Experimental results indicate that IAM is compatible with KV cache compression methods like $H_2O$ without compromising the model's ability of language modeling.

\begin{figure}[!htb]
\centering
\includegraphics[scale=0.2]{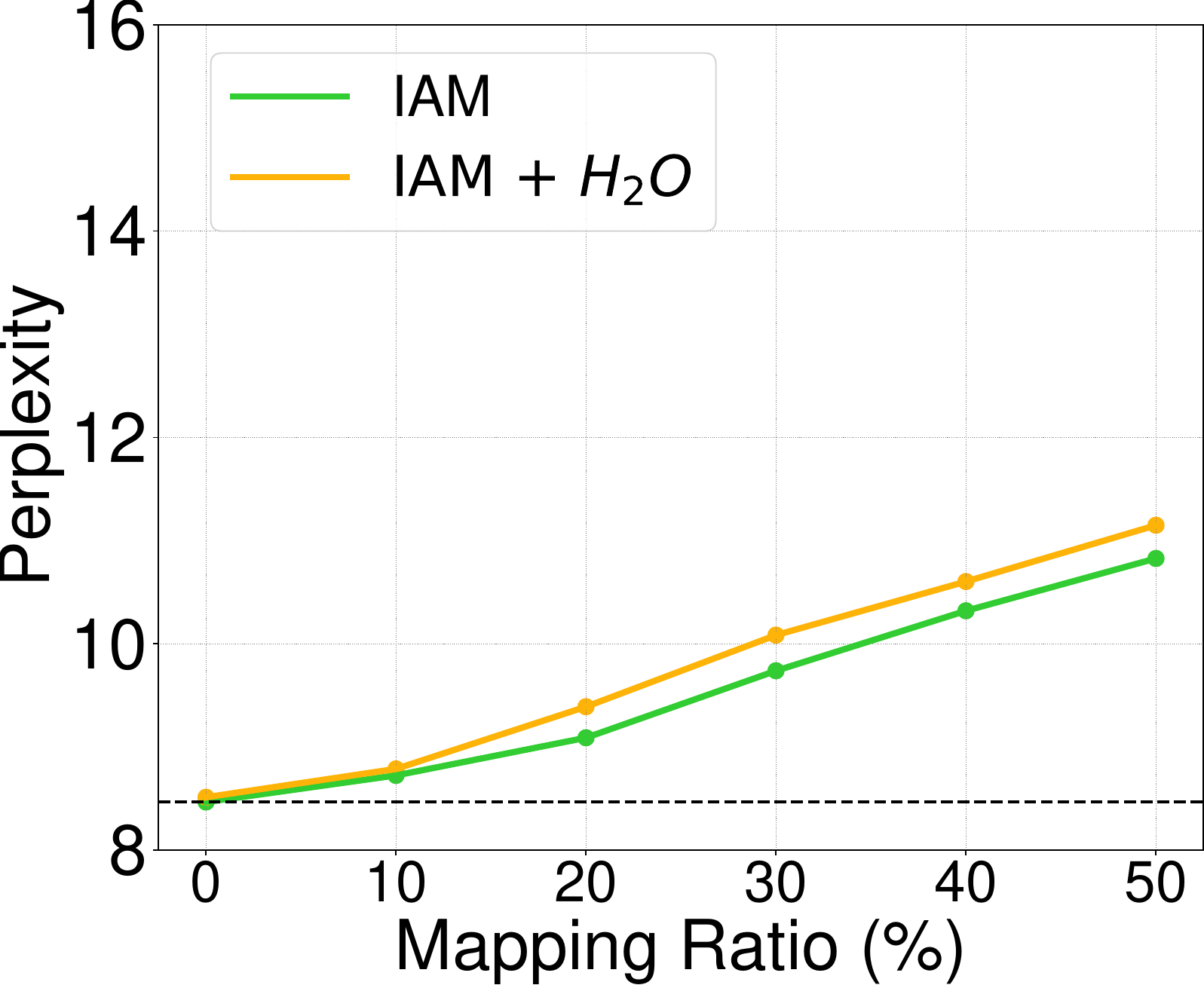}
\caption{The perplexity of using IAM in combination with KV cache compression method $H_2O$.}
\label{fig:layer_select_llama}
\end{figure}

\section{Conclusion}
We address the challenges faced by LLMs in serving long context scenarios by introducing IAM. This approach effectively reduces attention computation and KV cache usage by performing attention mapping between different-sized models with same series, with minimal impact on model performance. We also demonstrate that IAM is effective across different series of models. Furthermore, it is compatible with many existing KV cache optimization methods, making it a highly promising solution for deploying LLMs in resource-constrained environments.

\section*{Limitations}
There are also some limitations in our approach:  the current analysis and implementation of IAM are conducted at the granularity of model layers. If the granularity is further refined to the attention head, it could potentially yield additional improvements in model performance and resource savings. Additionally, IAM is not compatible with high-efficiency attention methods (e.g., Flash Attention) as it requires calculating attention scores. Besides that, IAM can not achieve total lossless for model performance.

% Bibliography entries for the entire Anthology, followed by custom entries
%\bibliography{anthology,custom}
% Custom bibliography entries only
\bibliography{custom}

\clearpage

\appendix

\section{Training Setting}
\label{app:train}
The model is trained using 8 NVIDIA A100 80GB GPUs. During training, only the parameters of the small model are updated through gradient descent. We set the batch size to 128 and train for a maximum of five epochs. The learning rate is set to 2e-5, with a weight decay of 0.01 applied during training. These training parameters remain consistent for both the Qwen series and Llama series models. We observe that the training loss stops decreasing significantly after two epochs; therefore, we adopt the small model trained after two epochs for subsequent evaluation experiments.

We compared the perplexity of Qwen2-72B using IAM on the WikiText-v2 dataset before and after instruction tuning in Table \ref{tab:train}. The results show that the perplexity decreases significantly after fine-tuning. Moreover, as the mapping ratio increases, the decrease becomes more pronounced. This indicates that appropriate fine-tuning at higher Mapping Ratios can help mitigate the performance degradation caused by IAM.

\begin{table}[h]
\resizebox{1\columnwidth}{!}{
\begin{tabular}{lccccc}
\toprule
\multicolumn{1}{c}{Mapping Ratio} & 10\%  & 20\%  & 30\%  & 40\%   & 50\%   \\ \midrule
w.o. tuning                       & 8.47 & 8.72 & 9.09 & 9.74  & 10.32 \\
w.  tuning                        & 8.53 & 8.84 & 9.18 & 10.34 & 11.99 \\ \bottomrule
\end{tabular}}
\caption{The ablation study on the impact of fine-tuning on IAM method.}
\label{tab:train}
\end{table}

\section{Extended Analysis of Attention Similarity}
\label{sec:appendix_1}

\paragraph{What is the degree of attention similarity between different-scale LLMs?}

\begin{figure*}[!h]
\centering
\includegraphics[scale=0.22]{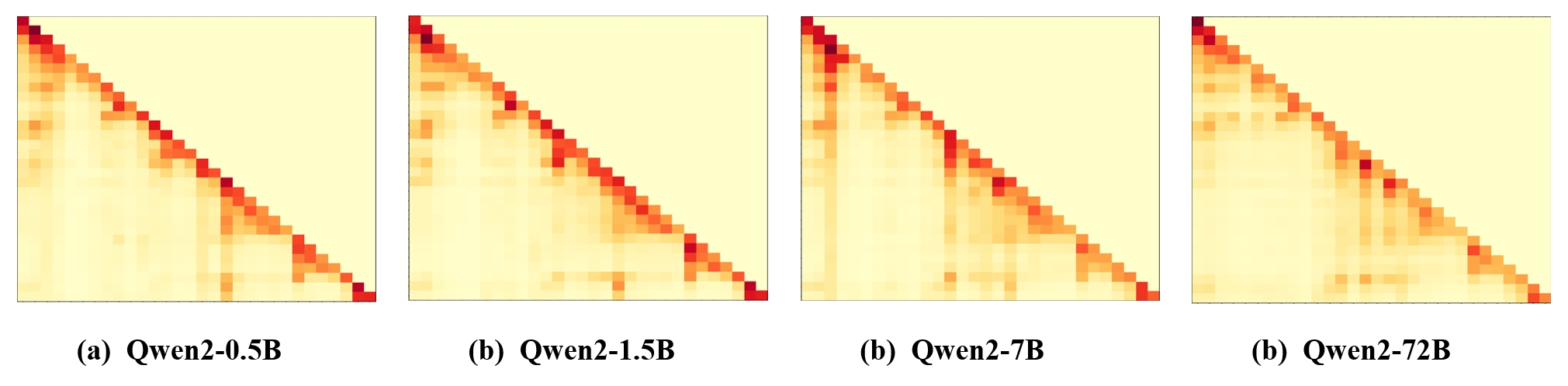}
\caption{The average attention matrices across all layers and heads from four different-scale models of Qwen2 series.}
\label{fig:attn_sim}
\end{figure*}

A fundamental premise of this paper is the high similarity of attention matrices across different-sized models within same series. We first demonstrate the visualization of average attention matrices across all layers and all heads under a given context in WikiText-v2 using four different-size models from the Qwen2 series, as shown in Figure \ref{fig:attn_sim}. It can be observed that these matrices exhibit strong similarities in their attention patterns.

We also provide a quantitative analysis of this similarity. Considering the WikiText-v2 dataset, we use the SLM of Qwen2-7B and the LLM of Qwen2-72B to compute the average cosine similarity of all pairwise most similar matrices. The result yield a average cosine similarity of 0.954. 

\paragraph{Why is such high similarity exhibited?}

The attention similarities between different-scale LLMs within same series are from our empirical observations. We speculate that it may be caused by the following factors: 1) They share the same model architecture, with differences only in parameters such as the number of layers and hidden dimensions. 2) They are pre-trained on the same corpus \cite{yang2024qwen2technicalreport}. 3) Some smaller models are distilled from larger models \cite{abdin2024phi3technicalreporthighly}. 4) The attention patterns in larger LLM exhibit more redundancy, especially in deeper decode layers, which means these layers tend to have simpler attention patterns for mapping by SLM.

\paragraph{Can different series models show the same similarity?}

Due to differences in training strategies and model architectures, the similarity between models from different series would drop significantly. For example, we test the case where the SLM is llama3.2-1B and the LLM is Qwen2-72B. Due to the difference of tokenizers, we employ the pairwise longest common subsequence method to align tokens between the two models. In this situation, the average cosine similarity is only 0.568. More critically, discrepancies in tokenizers result in attention matrices with different dimensions, which prevents direct attention mapping using IAM method.

\section{Detailed Analysis of Mapping Strategy}

\begin{figure}[!htb]
\centering
\includegraphics[scale=0.24]{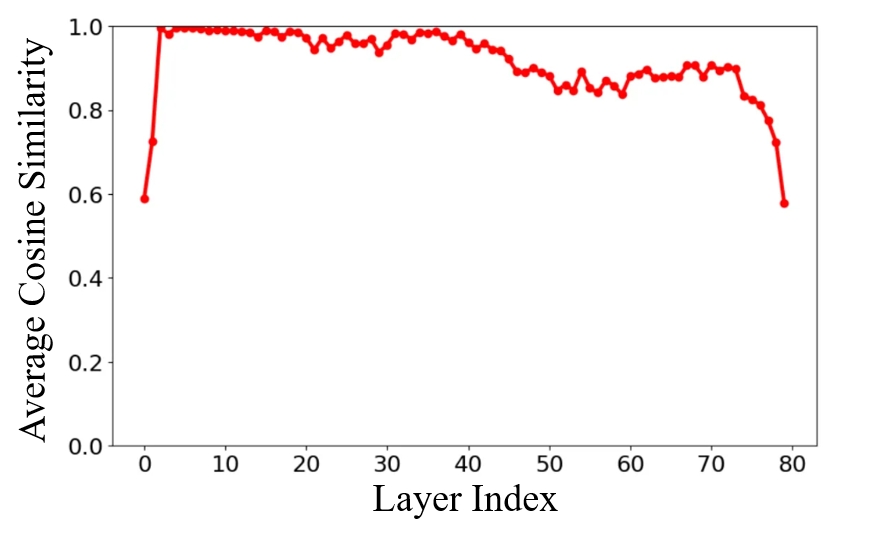}
\caption{The average cosine similarity for each layer of Qwen2-72B by calculating the most simular attention matrix between Qwen2-7B and Qwen2-72B.}
\label{fig:layer_sim}
\end{figure}

%In Section \ref{sec:layer_select}, we present a mapping strategy from a practical perspective, specifically by using benchmark scores to determine which layers are most suitable for mapping. 
In this section, we provide additional analysis of the mapping strategy. There is a natural consideration of mapping the layers with the highest average cosine similarity, aiming to minimize the loss introduced by mapping. To this end, we first calculate the average cosine similarity for each layer of Qwen2-72B, as shown in Figure \ref{fig:layer_sim}. It is evident that the layers in the first half of the model, excluding the first two layers, exhibit more strong similarity.

\begin{figure}[!htb]
\centering
\includegraphics[scale=0.26]{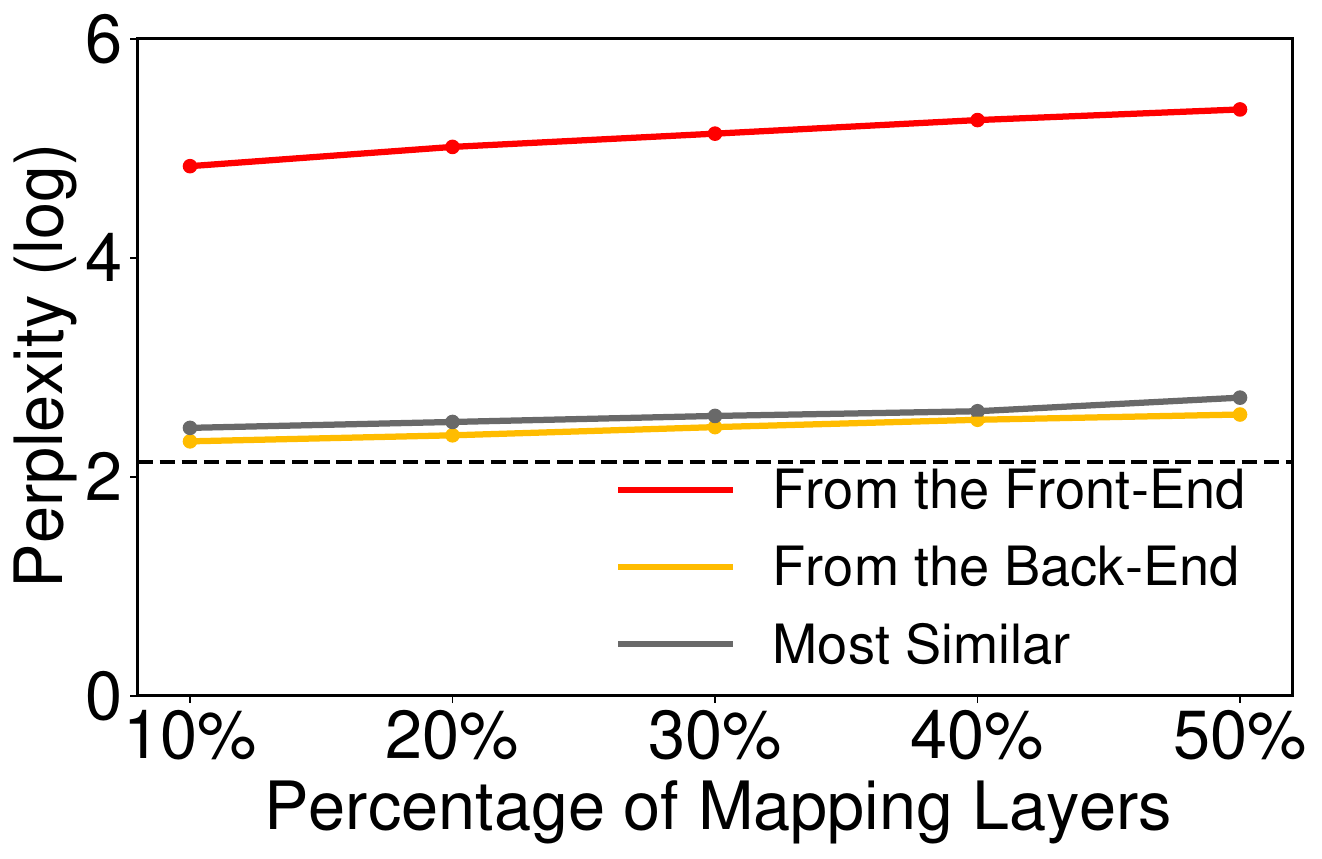}
\caption{The log perplexity of LLM according to different mapping layer selecting strategies.}
\label{fig:layer_select}
\end{figure}

We consider three mapping strategies: From the Front-End (mapping only the initial layers), From the Back-End (mapping only the latter layers), and the Most Similar (mapping the layers with the highest mean cosine similarity). Using the same evaluation methodology as described in section \ref{How to Measure Similarity}, the experimental results are illustrated in Figure \ref{fig:layer_select}, where the x-axis represents the percentage of layers mapped relative to the total number of layers. From experimental results, it is evident that the From the Front-End strategy is entirely unviable. The From the Back-End strategy yields the best results, even outperforming the Most Similar approach. We conjecture that this can be attributed to the cumulative error introduced by front-end mapping, which progressively accumulates through each subsequent layer. Consequently, introducing mappings at the back-end leads to fewer alterations from the original model, thereby maintaining a relatively greater capability of the LLM. This also explains why mapping the latter layers consistently yields better performance, both in the Qwen2 series models and the LLaMA3 series models.

\section{Statistical Analysis of Mapping Relations}

\begin{figure}[!htb]
\centering
\includegraphics[scale=0.15]{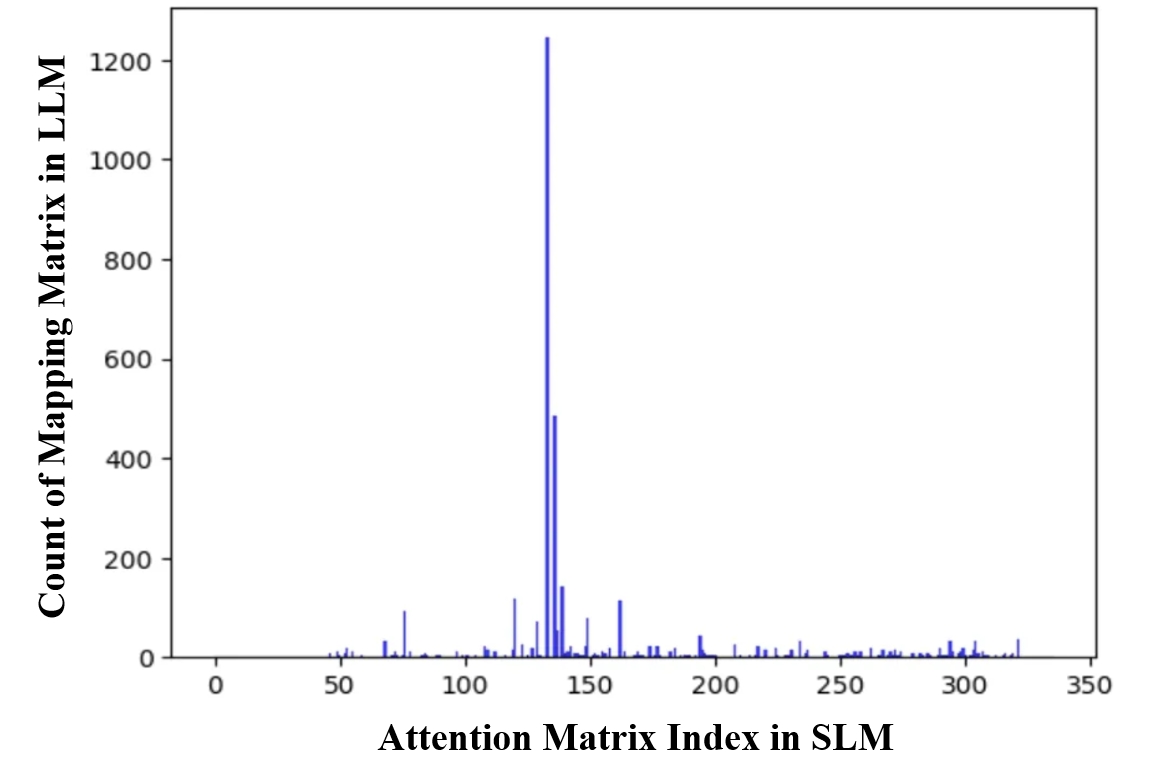}
\caption{The statistical analysis of the most frequent mapping relations between Qwen2-0.5B and Qwen2-72B.}
\label{fig:count}
\end{figure}

We also conduct a statistical analysis of the most frequent mapping relations on the WikiText-v2 dataset. Specifically, we record the mapping relations for each context. For each attention head of the LLM, we identify the mode of its mapping index. In this experiment, we used a SLM of Qwen2-0.5B and a LLM of Qwen2-72B, resulting in a mapping from 5120 to 360. The experimental results are shown in Figure \ref{fig:count}.

First, it can be observed that most matrices from the SLM are utilized in the mapping process, with the exception of the first forty matrices, which are rarely used. Additionally, the mapping relations exhibit a strong imbalance across different matrices. The most frequently used matrix is utilized more than 1200 times, significantly more than any other matrix. This finding also suggests that the attention patterns within the LLM have substantial sparsity. Such redundancy may partly explain why the LLM can extract key attention patterns from the more refined SLM, thereby maintaining performance without degradation.

\end{document}